\documentclass[11pt]{article}

\usepackage[a4paper,margin=1in]{geometry}
\usepackage[T1]{fontenc}
\usepackage[utf8]{inputenc}
\usepackage{lmodern}
\usepackage{microtype}
\usepackage{amsmath,amssymb,mathtools}
\usepackage{graphicx}
\usepackage[tableposition=top]{caption}
\usepackage{booktabs}
\usepackage{tabularx}
\usepackage{multirow}
\usepackage{xcolor}
\usepackage{enumitem}
\usepackage[numbers,sort&compress]{natbib}
\usepackage[hidelinks]{hyperref}
\usepackage{float}


\graphicspath{{figures/}}
\setlist{nosep}
\newcolumntype{Y}{>{\raggedright\arraybackslash}X}

\title{Overlap-Aware Sub-hypergraph Pattern Frequencies for Hypergraph-Level Classification}
\author{
  Nafiul Hasan Anik\\
  University of Dhaka\\
  {\normalsize\texttt{nafiulhasan-2019017811@cs.du.ac.bd}}
  \and
  Nazira Jesmin Lina\\
  University of Dhaka\\
  {\normalsize\texttt{nazirajesmin-2019117838@cs.du.ac.bd}}
  \end{tabular}\\[1em]\begin{tabular}[t]{c}
  Md Tanvir Alam\\
  University of Dhaka\\
  {\normalsize\texttt{tanvir@cse.du.ac.bd}}
  \and
  Md. Fahim Arefin\\
  University of Dhaka\\
  {\normalsize\texttt{fahim@cse.du.ac.bd}}
}
\date{}

\begin{document}

\maketitle

\begin{abstract}
Classifying whole hypergraphs requires a fixed-length description of inputs whose numbers of vertices and hyperedges vary. Frequent sub-hypergraph patterns provide such a description, but previous work records only whether each pattern is present, which loses two kinds of information: how large the overlap between two hyperedges is, and how much of a hypergraph each pattern makes up. We extend frequent sub-hypergraph mining with overlap labels, the binned intersection size of two pattern hyperedges inherited from the target hypergraph, and represent each hypergraph by the relative frequency of each mined pattern. The vocabulary is mined from training hypergraphs only and then frozen, and validation and test hypergraphs are matched against it. A support rule that depends on pattern size keeps the vocabulary small. On the six hypergraph-classification datasets of DHG-Bench, with a single configuration chosen on validation scores, the method has a higher mean Macro-F1 than the best published mean on five datasets, by 18.67 points on Twitter-Friend and 10.02 points on Steam-Player; on RHG-3 it is 0.21 points lower. Ablations show that on Steam-Player and Twitter-Friend, replacing relative frequencies with presence lowers Macro-F1 by about 18 points, and that removing overlap labels lowers it by 4 to 9 points.
\end{abstract}

\section{Introduction}
\label{sec:introduction}

A hypergraph extends a graph by allowing an edge, called a hyperedge, to join any number of vertices, so it can represent relations among groups of entities \citep{zhou2006learning}. In \emph{hypergraph-level classification}, each input is an entire hypergraph, and the goal is to assign one class label to it. For example, in the Steam-Player dataset each hypergraph describes one player. Its vertices are the games the player has played, and its hyperedges connect games that share tags. The task is to predict whether the player prefers single-player or multiplayer games \citep{feng2024hypergraph,li2025dhgbench}. In this work, we classify hypergraphs from their structure alone, that is, from the vertices, the hyperedges and the incidences between them.

Standard classifiers need a fixed-length input, but hypergraphs differ in their numbers of vertices and hyperedges. Hypergraph neural networks learn a fixed-length representation by aggregating vertex and hyperedge representations \citep{feng2019aaai-hypergraph,bai2021hypergraph,chien2022allset}. Frequent sub-hypergraph patterns provide a more explicit alternative. These patterns are small labeled sub-hypergraphs that occur in many hypergraphs of a collection \citep{alam2021mining}. When the data have no vertex or hyperedge attributes, vertices can be labeled by their degree and hyperedges by their cardinality, as in structure-only hypergraph kernels \citep{feng2024hypergraph}. Each mined pattern then serves as one feature, and previous work has represented a hypergraph by recording which patterns it contains \citep{alam2021mining,alam2023discovering}.

Presence features lose two kinds of information that matter in hypergraphs. The first is the size of an overlap. A pattern hyperedge may match part of a larger target hyperedge \citep{alam2021mining}, so a pattern records only a lower bound on how many vertices two hyperedges share. For example, a pattern whose two hyperedges share one vertex matches any two target hyperedges that share at least one. Two tags that share one game and two tags that share ten games therefore match the same patterns. Simple graphs have no such information, since two edges share at most one vertex. Hyperedge overlap has been studied as a structural property of real-world hypergraphs \citep{lee2021hyperedges}, and hypergraph motifs describe three connected hyperedges by which of their intersections are empty \citep{lee2020hypergraph}. The second is frequency. Presence records whether a pattern occurs, but not how much of the hypergraph it makes up. In dense hypergraphs many small patterns occur in almost every example, so their presence is nearly constant. Even where presence varies, two hypergraphs that contain the same patterns can differ in how often each occurs.

We address both losses within frequent sub-hypergraph mining. Each pattern vertex is labeled with the binned degree of the target vertex it matches. Each pattern hyperedge is labeled with the binned cardinality of the target hyperedge it matches, even when the pattern contains only part of that hyperedge. When two pattern hyperedges share a vertex, they also carry the binned size of the intersection of the two target hyperedges they match. Each hypergraph is represented by the number of times each pattern occurs in it, divided by the number of occurrences of the pattern's unlabeled form. These relative frequencies are passed to a standard neural classifier. The patterns are mined from the training hypergraphs. Every pattern of at most two incidences that occurs in at least two of them is kept, while larger patterns must occur in at least a fixed fraction of them (Section~\ref{sec:support}). This keeps the vocabulary bounded.

Building the vocabulary is itself a fitting step. If validation and test hypergraphs take part in mining, their structures affect which patterns are found and which pass the support threshold, even when their class labels are withheld. The procedure then becomes transductive at the mining stage. In our main protocol, the vocabulary is mined from training hypergraphs only and then frozen. Validation and test hypergraphs are matched against it without adding or changing patterns, and all scaling and model fitting use training hypergraphs only. A paired secondary study lets unlabeled validation and test hypergraphs take part in mining, while the classifier is still trained on training hypergraphs only.

We evaluate one configuration, chosen on validation scores alone, on the six hypergraph-classification datasets of DHG-Bench with their published partitions \citep{li2025dhgbench}. Its mean Macro-F1 is higher than the best mean reported in DHG-Bench on five datasets, with the largest margins on Twitter-Friend and Steam-Player. On IMDB-Dir-Form and RHG-10 the margin is within about one standard deviation, and on RHG-3 the method is slightly lower, with both close to 100\%. The DHG-Bench results are taken from the published paper and were not rerun. Against the hypergraph Weisfeiler--Lehman subtree kernel \citep{feng2024hypergraph}, run on the same partitions, the method's Macro-F1 is higher in every repeat on Steam-Player, Twitter-Friend and RHG-10, and slightly lower on RHG-3, where the kernel classifies every test hypergraph correctly. The ablations show that relative frequencies and overlap labels matter most on Steam-Player and Twitter-Friend, that patterns of three incidences help on the same two datasets at a higher mining cost, and that letting unlabeled held-out hypergraphs take part in mining gives no consistent benefit.

The contributions of this work are:
\begin{itemize}
    \item an overlap-aware extension of frequent sub-hypergraph mining, in which pairs of pattern hyperedges are labeled with the binned intersection size inherited from the target;
    \item a relative occurrence-frequency encoding of the mined patterns;
    \item a fully inductive evaluation on the six DHG-Bench hypergraph-classification datasets, in which the vocabulary is built from training hypergraphs only and one configuration is chosen on validation scores;
    \item descriptive comparisons with the published DHG-Bench results, and paired comparisons with hypergraph Weisfeiler--Lehman kernels run on the same partitions; and
    \item ablations of overlap labels, relative frequencies, label resolution, pattern size and vocabulary access, together with the mining cost of each pattern size.
\end{itemize}

\section{Related Work}
\label{sec:related}

\subsection{Hypergraph learning and hypergraph-level classification}

Learning on hypergraphs was first studied with spectral methods, which extend clustering, embedding and transductive classification from graphs to hypergraphs \citep{zhou2006learning}. Neural models followed. Hypergraph Neural Networks (HGNN) use a hyperedge convolution \citep{feng2019aaai-hypergraph}, and hypergraph convolution and attention (HCHA) operate on vertices and hyperedges together \citep{bai2021hypergraph}. HNHN applies nonlinear functions to both vertices and hyperedges \citep{dong2020hnhn}. AllSet, with instances such as AllSetTransformer, writes each layer as two learned multiset functions \citep{chien2022allset}. Other models turn a hypergraph into an ordinary graph before learning, for example by treating each vertex--hyperedge pair as a node (LEGCN) \citep{yang2022legcn} or by expanding vertices and hyperedges into one graph (HJRL) \citep{yan2024hjrl}. More recent models include DPHGNN \citep{saxena2024dphgnn} and the training-free TF-HNN \citep{tang2025tfhnn}. Several of these models were introduced for node classification. DHG-Bench compares them under a common protocol and includes hypergraph-level classification, which is the task studied here \citep{li2025dhgbench}.

Kernels offer a non-neural way to compare whole hypergraphs. The Weisfeiler--Lehman kernel for graphs with discrete labels compares graphs through repeated relabeling of their vertices \citep{shervashidze2011weisfeiler}. \citet{feng2024hypergraph} extend the Weisfeiler--Lehman test to hypergraphs and build two kernels on it. When no native vertex features are used, they initialize vertex labels from degrees and hyperedge labels from hyperedge degrees. Our method also labels vertices by degree and hyperedges by cardinality, but it represents a hypergraph by the counts of explicit mined patterns, not by the outcome of iterated relabeling.

\subsection{Frequent pattern mining for classification}

Frequent subgraph mining searches a collection of graphs for subgraphs that occur in many of them. gSpan avoids duplicate candidates by growing patterns depth-first in a canonical order \citep{yan2002gspan}, and CloseGraph mines only closed patterns to reduce the number of results \citep{yan2003closegraph}. Mined patterns have also been used as features for classification. For example, \citet{deshpande2005frequent} mine frequent substructures of chemical compounds and pass them to a classifier. For hypergraphs, \citet{horvath2007frequent} define frequent hypergraph mining and study its complexity. \citet{alam2021mining} allow a pattern hyperedge to match a subset of a target hyperedge, and mine such patterns with a depth-first search and canonical labeling. Their classification experiment records whether each pattern is present in a hypergraph. The journal version extends the framework to weighted and uncertain hypergraphs and repeats the classification experiment with presence features \citep{alam2023discovering}. For ordinary graphs, \citet{alam2021discriminating} mine in several phases with decreasing support thresholds until each graph is covered by enough patterns, and learn a supervised graph embedding from them.

We build on the containment and mining framework of \citet{alam2021mining}. Our method differs from theirs in the labels (binned degree, cardinality and overlap size), the relative frequencies used in place of presence, the support rule that depends on pattern size, and the vocabulary, which is mined from training hypergraphs only.

\subsection{Counting small configurations}

Counting small substructures is a common way to describe a graph. Graphlet kernels compare graphs through the counts of their subgraphs on three to five vertices \citep{shervashidze2009efficient}. For hypergraphs, hypergraph motifs describe three connected hyperedges by the connectivity pattern of their intersections, and their counts form a profile of a hypergraph \citep{lee2020hypergraph}. Hyperedge overlap has also been measured directly in real-world hypergraphs \citep{lee2021hyperedges}. Our overlap labels are defined on pairs of hyperedges, and the counted patterns are mined from the data, not taken from a fixed motif set.

\section{Preliminaries and Problem Definition}
\label{sec:preliminaries}

A hypergraph $H=(V,E)$ has a set of vertices $V$ and a set of hyperedges $E$, each of which contains a nonempty subset of $V$. Two distinct hyperedges may contain the same vertices. A vertex $v$ and a hyperedge $e$ form an \emph{incidence} when $v\in e$. The degree $d(v)$ is the number of hyperedges that contain $v$, and the cardinality $|e|$ is the number of vertices in $e$. We replace each exact value $x$ by the logarithmic bin $b(x)$, where $b(x)=0$ if $x=0$ and $b(x)=\lceil\log_2 x\rceil+1$ otherwise. Thus the values $1$, $2$, $3$--$4$, $5$--$8$ and so on fall into bins $1$, $2$, $3$, $4$ and so on. A vertex $v$ is labeled $b(d(v))$ and a hyperedge $e$ is labeled $b(|e|)$. Two hyperedges $e$ and $f$ that share at least one vertex also receive an \emph{overlap label} $b(|e\cap f|)$. All labels are computed inside a single hypergraph from its structure alone and use no class labels.

A \emph{pattern} is a small connected hypergraph with vertex labels, hyperedge labels, and an overlap label for every pair of its hyperedges that share a vertex. Its \emph{size} is its number of incidences, and its \emph{skeleton} is the same pattern with all labels removed. Mined patterns have at most $D$ incidences, where the maximum size $D$ is fixed in advance (Section~\ref{sec:setup}). A pattern $p$ \emph{occurs} in a hypergraph $H$ if its vertices and hyperedges can be mapped to distinct vertices and distinct hyperedges of $H$ so that (i) every mapped vertex and hyperedge has the same label as its image; (ii) every vertex of a pattern hyperedge is mapped into the image of that hyperedge; and (iii) every pair of pattern hyperedges that share a vertex is mapped to a pair of hyperedges with the same overlap label. The image of a hyperedge may contain further vertices that the pattern does not use, so a pattern may hold only part of a hyperedge. Pattern labels are inherited from the target. For example, a pattern that holds two vertices of a three-vertex hyperedge carries the hyperedge label $b(3)$, not $b(2)$. Similarly, two pattern hyperedges that share one pattern vertex carry the overlap label of the target hyperedges they match, which may share many more vertices. Pairs of pattern hyperedges that share no pattern vertex are not constrained.

The \emph{occurrence count} $c(p,H)$ is the number of distinct images of $p$ in $H$. An image is the set of incidences of $H$ used by a match, so matches that differ only by a symmetry of the pattern are counted once. Counts are computed for patterns of at most $D$ incidences. Let $c(\bar p,H)$ be the number of occurrences in $H$ of the skeleton $\bar p$ of $p$, over all labelings, including those that are not in the vocabulary. The \emph{relative frequency} of $p$ in $H$ is
\[
r(p,H)=\frac{c(p,H)}{c(\bar p,H)},
\]
and is set to $0$ when $c(\bar p,H)=0$. The \emph{support} of $p$ in a set of hypergraphs is the number of those hypergraphs in which $p$ occurs at least once, so it counts hypergraphs, not matches.

\begin{figure}[t]
  \centering
  \includegraphics[width=\linewidth]{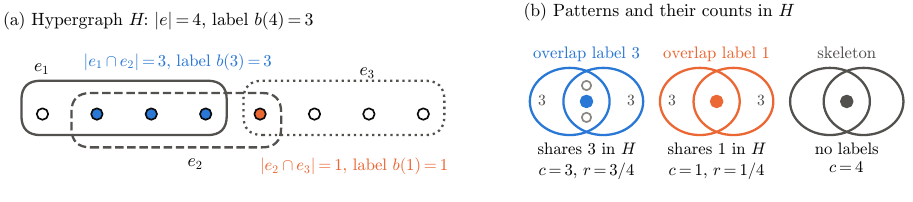}
  \caption{Overlap labels and relative frequencies in an example hypergraph. (a) Hypergraph $H$, whose hyperedges all have four vertices (label $b(4)=3$). The colored vertices have degree two (label $b(2)=2$), and the others have degree one. (b) Two patterns of size two that differ only in their overlap label, and their common skeleton, with their counts $c$ and relative frequencies $r$ in $H$. Each pattern hyperedge holds one vertex of a hyperedge of $H$ and inherits its labels. Blue and orange mark where each pattern occurs in $H$. The faded vertices are the further vertices shared by $e_1$ and $e_2$, and are not part of the pattern.}
  \label{fig:example}
\end{figure}

Figure~\ref{fig:example} illustrates these definitions on an example hypergraph. In Figure~\ref{fig:example}(a), hyperedges $e_1$ and $e_2$ share three vertices, and $e_2$ and $e_3$ share one. Consider a pattern of size two, with two hyperedges at a common vertex. It holds only one vertex of each target hyperedge, so without overlap labels both pairs match the same pattern, which occurs four times and has relative frequency~$1$. With overlap labels, it splits into the two patterns of Figure~\ref{fig:example}(b), with overlap labels $b(3)=3$ and $b(1)=1$. The first occurs three times, once at each vertex shared by $e_1$ and $e_2$, and the second once. Their common skeleton occurs four times, so their relative frequencies are $3/4$ and $1/4$.

Let $\{(H_i,y_i)\}_{i=1}^{N}$ be a collection of hypergraphs with class labels $y_i$. The task is to predict the class of a hypergraph that was not used for fitting. In each repeat, the indices are split into disjoint training, validation and test sets. In the main, fully inductive protocol, patterns are mined from the training hypergraphs only, and the resulting vocabulary is frozen. Validation and test hypergraphs are only matched against it. The scaling of the features and the classifier are fitted on the training hypergraphs, and the validation set guides model selection. In the secondary transductive-vocabulary protocol, the unlabeled validation and test hypergraphs also take part in mining, while the scaling and the classifier are still fitted on the training hypergraphs only.

\section{Method}
\label{sec:method}

\subsection{Overview}
\label{sec:overview}

\begin{figure}[t]
  \centering
  \includegraphics[width=\linewidth]{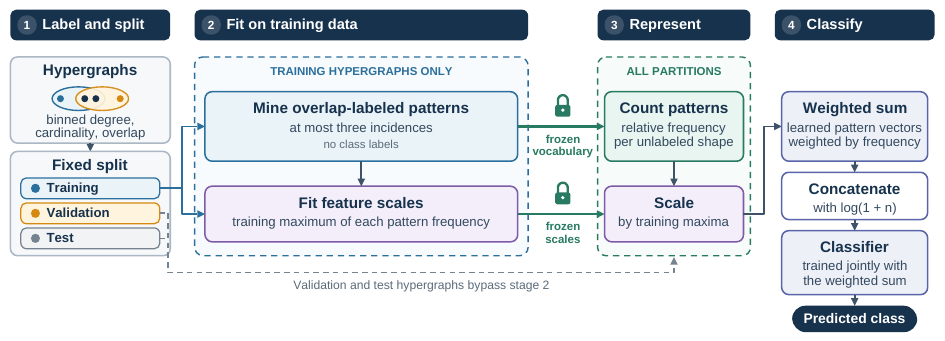}
  \caption{The fully inductive pipeline. Hypergraphs are labeled from their structure and split. Overlap-labeled patterns and feature scales are fitted on the training hypergraphs only and then frozen. Counting, scaling and classification are applied to all partitions, so validation and test hypergraphs only pass through the frozen vocabulary and scales.}
  \label{fig:overview}
\end{figure}

Figure~\ref{fig:overview} shows the pipeline. Every hypergraph is labeled from its own structure (Section~\ref{sec:preliminaries}) and the indices are split into training, validation and test sets. Patterns are then mined from the training hypergraphs only, without class labels, which gives a vocabulary of patterns of at most $D$ incidences. The vocabulary is frozen. Every hypergraph, whatever its partition, is turned into a vector of relative frequencies over this vocabulary. The scale of each feature is the maximum over the training hypergraphs, and it is frozen as well. A classifier trained on the scaled training vectors predicts the class. The vocabulary, the scales and the classifier weights are fitted on the training hypergraphs only. The validation loss only guides model selection (Section~\ref{sec:classifier}), and test hypergraphs are used only for the final evaluation.

\subsection{Overlap-aware pattern growth}
\label{sec:growth}

We mine with the frequent sub-hypergraph miner of Alam et al.~\citep{alam2021mining}, which grows a pattern one incidence at a time in a depth-first search. An extension either adds a new hyperedge at a vertex that is already in the pattern, or adds a vertex to the hyperedge that was added last. Each pattern has a canonical code, which gives it a unique extension sequence. A candidate that is not in this minimal form is rejected, so no pattern is generated twice.

We add overlap labels to this search. When an extension makes a new hyperedge share a vertex with earlier hyperedges for the first time, it also records the overlap label of each such pair, taken from the two target hyperedges they match. From size three on, a hyperedge added at a vertex can join several earlier hyperedges at once, and it then records one label for each of them. The candidate labels are taken from the pairs of target hyperedges that exist at that vertex. Each label is fixed at that point, because a pair of hyperedges becomes connected in the pattern exactly once. The canonical order uses the overlap labels only to break ties between extensions that are otherwise equal. Matching checks them as condition (iii) of Section~\ref{sec:preliminaries}. The miner counts a candidate once for each training hypergraph in which it occurs, which is the support defined above.

\subsection{Size-dependent support}
\label{sec:support}

Let $\mathcal{T}$ be the set of training hypergraphs and $\rho\in(0,1)$ a fixed fraction. A candidate $p$ of size $|p|\le D$ is kept when
\[
\operatorname{supp}(p,\mathcal{T}) \;\ge\;
\begin{cases}
2 & \text{if } |p|\le 2,\\
\lceil \rho\,|\mathcal{T}| \rceil & \text{if } 2<|p|\le D,
\end{cases}
\]
where $\operatorname{supp}(p,\mathcal{T})$ is the support of $p$ in $\mathcal{T}$. Patterns of one or two incidences are few, so keeping all that recur in at least two training hypergraphs adds little to the vocabulary. Larger patterns are far more numerous, and the fraction $\rho$ bounds their number. The values of $\rho$ and $D$ are given in Section~\ref{sec:setup}. Support is computed on the training hypergraphs and uses no class labels. A candidate that fails its threshold is not extended. This pruning loses no pattern, because support cannot increase when a pattern grows. The search records every candidate that passes before extending it, so the vocabulary holds the kept patterns of every size up to $D$, and not only those of size $D$.

For graphs, \citet{alam2021discriminating} lower the support threshold in phases until every training graph is covered by enough patterns, with a fallback for graphs that remain uncovered. We apply this procedure to hypergraphs, with a target of one pattern per training hypergraph. With the rule above, however, every training hypergraph contains a kept pattern after the first phase on all six datasets. The later phases and the fallback therefore never add a pattern, and the rule above is the one that takes effect.

\subsection{Occurrence counts and relative frequency}
\label{sec:counting}

For a hypergraph $H$ and a vocabulary pattern $p$ we compute the relative frequency $r(p,H)$ of Section~\ref{sec:preliminaries}. Dividing by the skeleton count reduces the dependence on the size of $H$. A large hypergraph has more occurrences of every pattern, but the share of a pattern among the occurrences of its skeleton is comparable between hypergraphs of different sizes. A hypergraph that has no occurrence of $p$ has $r(p,H)=0$, so $r(p,H)>0$ exactly when $p$ occurs in $H$.

The counts are obtained for all hypergraphs in one pass that tallies the occurrences of each connected skeleton of at most $D$ incidences. The tallies use no class labels and involve no fitting, so a validation or test hypergraph gets its counts from the frozen vocabulary in the same way as a training hypergraph. As a check, the pipeline also runs the miner's matcher with the frozen vocabulary. For every pattern and every hypergraph, it verifies that the count is positive exactly when the matcher reports that the pattern occurs.

\subsection{Classifier}
\label{sec:classifier}

The classifier is a small neural network built from standard components. The vocabulary has $M$ patterns. For the $i$-th hypergraph, the feature vector $x_i\in\mathbb{R}^{M}$ holds the relative frequencies divided by the scale of each feature,
\[
x_{ij}=\frac{r(p_j,H_i)}{s_j},\qquad s_j=\max_{H\in\mathcal{T}} r(p_j,H),
\]
where $s_j$ is set to $1$ if the maximum is $0$. The scale is fitted on the training hypergraphs and is not centered, so an absent pattern stays exactly zero.

A bias-free linear layer with weight matrix $W\in\mathbb{R}^{M\times d}$ maps $x_i$ to a $d$-dimensional vector. Its output is a weighted sum of learned pattern vectors, with the scaled frequencies as weights. The vector is layer-normalized and concatenated with $\log(1+n_i)$, where $n_i$ is the number of patterns present in $H_i$:
\[
z_i=\bigl[\,\operatorname{LN}(W^{\top}x_i)\;;\;\log(1+n_i)\,\bigr].
\]
During training, a fraction of the entries of $x_i$ is dropped at random before the linear layer. The count $n_i$ is always computed from the original, undropped vector. The vector $z_i$ goes through a multilayer perceptron with two hidden layers, each followed by ReLU, layer normalization and dropout, and a softmax output. The weights are trained jointly with Adam on the cross-entropy loss with $L_2$ regularization. The validation loss is used only for early stopping and learning-rate scheduling. Section~\ref{sec:setup} gives the remaining hyperparameters.

\subsection{Data access and cost}
\label{sec:cost}

In the fully inductive protocol, only training hypergraphs are used for mining, the vocabulary rule, the feature scales and the classifier weights. Validation and test hypergraphs are matched against the frozen vocabulary and passed through the frozen scales. They take no part in these steps, except that the validation loss is used for early stopping and learning-rate scheduling. The vocabulary, the scales and the model are fitted again in every repeat.

In the transductive-vocabulary protocol, the unlabeled validation and test hypergraphs are also mined. The support rule is then applied to all hypergraphs, with $|\mathcal{T}|$ replaced by the number of hypergraphs in the collection. The scales and the classifier still use the training hypergraphs only, and only the patterns that occur in at least two training hypergraphs enter the classifier. A pattern that is frequent only in the held-out hypergraphs therefore has no weight in the model and is dropped. The protocol thus changes the vocabulary and nothing else.

Mining is the expensive step. It enumerates candidate extensions and matches each one against the supporting hypergraphs, and its cost grows with the maximum pattern size (Section~\ref{sec:results}). Counting is a single tally per hypergraph and does not search for each pattern separately. The miner's matcher is run on the validation and test hypergraphs only for the check of Section~\ref{sec:counting}. The classifier input is a dense matrix with one row per hypergraph and one column per pattern, so its storage grows with $N\times M$.

\section{Experimental Setup}
\label{sec:setup}

\subsection{Research questions}
\label{sec:rqs}

The experiments address five research questions:

\begin{description}[leftmargin=3.2em,style=nextline]
    \item[RQ1] How does the method compare with the published DHG-Bench results and with hypergraph Weisfeiler--Lehman kernels run on the same partitions?
    \item[RQ2] How much do overlap labels and relative frequencies contribute?
    \item[RQ3] How do classification scores, vocabulary size and mining time change with the maximum pattern size?
    \item[RQ4] How sensitive are the results to the label resolution, the support rule, the patterns of one incidence, pattern selection and the components of the classifier?
    \item[RQ5] Does letting the unlabeled validation and test hypergraphs take part in mining change the results?
\end{description}

\subsection{Datasets}
\label{sec:datasets}

We use the six hypergraph-classification datasets of DHG-Bench \citep{li2025dhgbench}, which were introduced by \citet{feng2024hypergraph}. Four are real-world datasets and two are synthetic. IMDB-Dir-Form and IMDB-Dir-Genre are built from co-director relationships in IMDB, with one hypergraph per film. They classify each film by its form (for example, animation or documentary) or by its genre. Steam-Player was described in Section~\ref{sec:introduction}. In Twitter-Friend, each hypergraph is the friendship network of one Twitter user, and the class records whether the user posted about ``National Dog Day'' or ``Respect Tyler Joseph''. RHG-10 and RHG-3 contain randomly generated hypergraphs built from ten and three structure types, respectively. RHG-10 includes three pairs of types that are easily confused, and RHG-3 uses hyper-pyramids, hyper-check-tables and hyper-wheels \citep{feng2024hypergraph}.

Our input files contain the same hypergraphs as the DHG-Bench data files, in the same order and with the same class labels. Repeated hyperedges are kept as separate hyperedges, as in those files. \citet{feng2024hypergraph} merge them, so their average numbers of hyperedges for IMDB-Dir-Form, IMDB-Dir-Genre and Twitter-Friend are slightly lower than those in Table~\ref{tab:datasets}. The method reads only vertices, hyperedges and incidences. It uses no native attributes or entity identities. Table~\ref{tab:datasets} also shows how often hyperedges that meet share more than one vertex. On Steam-Player and Twitter-Friend, two hyperedges that meet at a vertex also share another vertex in 79.2\% and 73.5\% of the cases, respectively. On the other four datasets, this share ranges from 14.8\% (RHG-10) to 63.5\% (RHG-3). Because a pair is counted once per shared vertex, pairs with large overlaps carry more weight. In terms of patterns, the share is the fraction of occurrences of the size-two pattern (two hyperedges at one vertex) whose overlap label is at least $b(2)$.

\begin{table}[t]
  \centering
  \caption{Datasets. Train/val./test gives the partition sizes, which are the same in all five repeats. Avg.\ $|E|$ counts repeated hyperedges separately. Largest class is the share of hypergraphs in the most frequent class. Overlap $\ge 2$: over all cases in which two hyperedges meet at a vertex, so that a pair is counted once per shared vertex, the share in which the two hyperedges also share another vertex.}
  \label{tab:datasets}
  \small
  \setlength{\tabcolsep}{3.5pt}
  \begin{tabular}{llrrcrrrr}
    \toprule
    Dataset & Type & \begin{tabular}[b]{@{}r@{}}Hyper-\\graphs\end{tabular} & Classes & Train/val./test & Avg.\ $|V|$ & Avg.\ $|E|$ & \begin{tabular}[b]{@{}r@{}}Largest\\class (\%)\end{tabular} & \begin{tabular}[b]{@{}r@{}}Overlap\\$\ge 2$ (\%)\end{tabular} \\
    \midrule
    IMDB-Dir-Form  & Real      & 1,869 & 3  & 1,495/186/188 & 15.72 & 39.30 & 42.7 & 40.5 \\
    IMDB-Dir-Genre & Real      & 3,393 & 3  & 2,714/339/340 & 17.33 & 36.55 & 51.3 & 41.6 \\
    Steam-Player   & Real      & 2,048 & 2  & 1,638/204/206 & 13.83 & 46.36 & 50.0 & 79.2 \\
    Twitter-Friend & Real      & 1,310 & 2  & 1,048/131/131 & 21.65 & 84.77 & 59.2 & 73.5 \\
    RHG-10         & Synthetic & 2,000 & 10 & 1,600/200/200 & 31.32 & 29.77 & 10.8 & 14.8 \\
    RHG-3          & Synthetic & 1,500 & 3  & 1,200/150/150 & 35.49 & 17.95 & 33.5 & 63.5 \\
    \bottomrule
  \end{tabular}
\end{table}

\subsection{Evaluation protocol}
\label{sec:protocol}

We use the DHG-Bench partitions \citep{li2025dhgbench}, regenerated with the split procedure of its released code. For repeat $r=0,\dots,4$, PyTorch's \texttt{random\_split} with seed $42+r$ assigns 80\% of the hypergraphs to training, 10\% to validation and the rest to testing, with the training and validation sizes rounded down. Because the hypergraph order matches the DHG-Bench files, each repeat has the same training, validation and test hypergraphs as in DHG-Bench. The partitions are unstratified repeated holdout splits, not cross-validation folds. Every method that we run uses the same saved partitions.

All six datasets share one configuration. The search covered whether overlap labels are used, the maximum pattern size $D$ (two or three), the support fraction $\rho$ for patterns of three incidences (0.2 or 0.1), the encoding of the counts (relative frequency or $\log(1+c(p,H))$) and the strength of the classifier's regularization (two levels). We chose the configuration with the highest mean validation Macro-F1 over the six datasets and five repeats, with mean validation accuracy as the tie-breaker. This rule was fixed in advance, and test scores were not used for any choice. Larger values of $D$ were left out of the search, because mining time grows quickly with pattern size. From $D=2$ to $D=3$, it already rises by a factor of 11 to 24 on the real-world datasets (Figure~\ref{fig:depth}). We evaluate $D=4$ as one of the ablations (Section~\ref{sec:rq3}). Each ablation changes one component of the selected configuration and is reported regardless of its outcome. No ablation was used to revise the configuration. Within each training run, the validation loss controls the learning rate, early stopping and the choice of checkpoint (Section~\ref{sec:classifier}).

The transductive-vocabulary study (Section~\ref{sec:cost}) uses the same partitions, seeds and settings as the main protocol, so the vocabulary is the only input that differs. We report it separately from the main results.

\subsection{Baselines, metrics and paired comparisons}
\label{sec:baselines}

\paragraph{Published results.} We take the published results from Table~A7 of DHG-Bench \citep{li2025dhgbench}, as given in its arXiv version~2 (arXiv:2508.12244v2). It reports fifteen hypergraph neural networks and three simpler models, which are an MLP and a graph convolutional network and a graph attention network on the clique expansion. For each dataset and metric we report the highest published mean, so the reference model can differ between Macro-F1 and accuracy. In DHG-Bench's released code, the models receive a one-hot encoding of each vertex's exact degree as vertex features, so they also use structure alone. The test hypergraphs are the same as ours, but training and model selection follow DHG-Bench, so the comparison does not isolate the effect of the representation.

\paragraph{Hypergraph Weisfeiler--Lehman kernels.} We also run the hypergraph Weisfeiler--Lehman subtree and hyperedge kernels of \citet{feng2024hypergraph} on our partitions. We use the default settings of the authors' code, which are four iterations, cosine normalization and a support vector machine with $C=1$. The colors are learned from the training hypergraphs only. Validation and test hypergraphs are compared with the training hypergraphs, and, as in the authors' code, colors that never occur in training are ignored. The main variant is the subtree kernel on the authors' input, with exact degree and cardinality labels and repeated hyperedges merged. Appendix~\ref{app:baselines} reports the hyperedge kernel and the variants with our log2 labels. Our implementation agrees with the authors' code up to floating-point precision on all six datasets.

\paragraph{Metrics.} Macro-F1 is the primary metric and accuracy the secondary one. Macro-F1 averages the F1 scores of all classes with equal weight, and a class whose F1 score is undefined counts as zero. Scores are percentages, reported as the mean and population standard deviation over the five repeats. DHG-Bench standard deviations are reproduced as published. As a reference point, we also assign every held-out hypergraph the most frequent class among the training hypergraphs of its repeat. On Twitter-Friend, whose largest class holds 59.2\% of the hypergraphs, this gives 58.78\% accuracy but 37.00\% Macro-F1.

\paragraph{Paired comparisons.} All variants that we run share the five partitions, so we compare them repeat by repeat. Each difference is a variant's score minus the full model's, on validation and on test. In the vocabulary-access study, it is the transductive score minus the inductive one. We report the mean difference in points and the number of repeats in which it is positive or negative, counting differences below 0.005 points in absolute value as ties. With only five repeated partitions, these counts do not establish statistical significance, and a small difference does not show that two variants are equivalent. The published DHG-Bench results are aggregates, so they cannot be compared repeat by repeat.

\subsection{Implementation details}
\label{sec:implementation}

The pipeline is written in Python. The classifier uses TensorFlow/Keras \citep{abadi2016tensorflow}, and the kernel baseline uses scikit-learn. All experiments ran on the CPU of one machine (Intel Core i7-13700 with 16 cores and 24 hardware threads, 32\,GB of RAM) under Ubuntu~24.04. The software versions are Python~3.12.3, TensorFlow~2.21.0 and scikit-learn~1.8.0. Each repeat has a fixed model seed, and TensorFlow's deterministic operations are enabled. A rerun on the same hardware and software, with the same thread settings, gives the same results. The six datasets were run concurrently for both mining and classification, with TensorFlow's default thread settings. The reported times therefore include CPU sharing between runs. Mining results are cached by configuration. When a cache is reused, the reported mining time is the one recorded when the cache was built.

Degree, cardinality and overlap labels use the log2 bins of Section~\ref{sec:preliminaries}. The selected maximum pattern size is $D=3$. Patterns of one or two incidences must occur in at least two training hypergraphs. Patterns of three incidences must occur in at least 10\% of them ($\rho=0.1$), with the threshold rounded up. The encoder has 32 dimensions, a pattern dropout rate of 0.3 and an $L_2$ coefficient of 0.05. The two hidden layers have 64 and 32 units, dropout rates of 0.3 and 0.2 and $L_2$ coefficients of 0.05 and 0.03. The weaker regularization level has no pattern dropout, an encoder $L_2$ coefficient of 0.01 and hidden-layer $L_2$ coefficients of 0.01 and 0.005. Training uses Adam \citep{kingma2015adam} with an initial learning rate of $3\times10^{-4}$, batch size 16 and at most 1,000 epochs. The learning rate is halved after five epochs without improvement in validation loss, with a minimum of $10^{-5}$. Training stops after 20 epochs without improvement. The learning rate, batch size and patience were not varied in any experiment. The same settings are used for all datasets and in both vocabulary-access protocols.

\section{Results}
\label{sec:results}

\subsection{Comparison with published results and hypergraph kernels (RQ1)}
\label{sec:rq1}

Table~\ref{tab:main-results} compares the test scores of the full model with the best published DHG-Bench result for each dataset. With one configuration for all six datasets, the method has a higher mean Macro-F1 than the best published mean on five datasets. The margin is largest on Twitter-Friend (18.67 points) and Steam-Player (10.02 points), followed by IMDB-Dir-Genre (3.09 points). On IMDB-Dir-Form (1.71 points) and RHG-10 (0.91 points), it is within about one standard deviation of either result. On RHG-3 the method is 0.21 points lower, with both results above 99.7\%. Accuracy gives the same picture, with margins of 14.20 points on Twitter-Friend and 8.74 points on Steam-Player. On IMDB-Dir-Form the two means differ by 0.11 points, and on RHG-3 the method is 0.19 points lower.

\begin{table}[t]
  \centering
  \caption{Comparison with published results. Test scores (\%), mean $\pm$ population standard deviation over five repeats. DHG-Bench: the highest published mean per dataset and metric (Table~A7 of \citet{li2025dhgbench}) and the model that reports it. The published values are not rerun and are not paired with our runs.}
  \label{tab:main-results}
  \small
  \begin{tabular}{lccl}
    \toprule
    & & \multicolumn{2}{c}{DHG-Bench (published)} \\
    \cmidrule(lr){3-4}
    Dataset & Ours & Best mean & Model \\
    \midrule
    \multicolumn{4}{l}{\emph{Macro-F1}} \\
    IMDB-Dir-Form & 59.63 $\pm$ 1.91 & 57.92 $\pm$ 2.24 & HGNN~\citep{feng2019aaai-hypergraph} \\
    IMDB-Dir-Genre & 76.82 $\pm$ 1.37 & 73.73 $\pm$ 1.92 & HJRL~\citep{yan2024hjrl} \\
    Steam-Player & 62.74 $\pm$ 1.51 & 52.72 $\pm$ 2.54 & TF-HNN~\citep{tang2025tfhnn} \\
    Twitter-Friend & 72.05 $\pm$ 3.68 & 53.38 $\pm$ 4.92 & LEGCN~\citep{yang2022legcn} \\
    RHG-10 & 98.17 $\pm$ 0.83 & 97.26 $\pm$ 1.04 & AllSetTransformer~\citep{chien2022allset} \\
    RHG-3 & 99.74 $\pm$ 0.32 & 99.95 $\pm$ 0.02 & HNHN~\citep{dong2020hnhn} \\
    \midrule
    \multicolumn{4}{l}{\emph{Accuracy}} \\
    IMDB-Dir-Form & 64.15 $\pm$ 2.01 & 64.04 $\pm$ 2.70 & DPHGNN~\citep{saxena2024dphgnn} \\
    IMDB-Dir-Genre & 80.71 $\pm$ 1.05 & 78.12 $\pm$ 1.96 & HCHA~\citep{bai2021hypergraph} \\
    Steam-Player & 63.59 $\pm$ 1.19 & 54.85 $\pm$ 1.82 & TF-HNN~\citep{tang2025tfhnn} \\
    Twitter-Friend & 74.35 $\pm$ 3.63 & 60.15 $\pm$ 1.70 & AllSetTransformer~\citep{chien2022allset} \\
    RHG-10 & 98.20 $\pm$ 0.87 & 97.30 $\pm$ 0.98 & AllSetTransformer~\citep{chien2022allset} \\
    RHG-3 & 99.73 $\pm$ 0.33 & 99.92 $\pm$ 0.02 & HNHN~\citep{dong2020hnhn} \\
    \bottomrule
  \end{tabular}
\end{table}

Table~\ref{tab:hgwl} compares the method with the hypergraph Weisfeiler--Lehman subtree kernel. Because the kernel runs on our partitions, the two can be compared repeat by repeat. The method's Macro-F1 is higher in all five repeats on Steam-Player (by 19.13 points on average), Twitter-Friend (17.52 points) and RHG-10 (1.85 points), and in four of the five on IMDB-Dir-Genre (2.42 points). On IMDB-Dir-Form, Macro-F1 is 0.78 points higher on average, with three repeats higher and two lower. Accuracy on this dataset, however, is 2.45 points lower on average and lower in all five repeats. On RHG-3, the kernel classifies every test hypergraph correctly. The method is 0.26 points lower on average, with two repeats lower and three equal. In Macro-F1, the subtree kernel is within about one standard deviation of the best published mean, or above it, on five datasets, and 9.12 points below on Steam-Player. Appendix~\ref{app:baselines} reports the other kernel variants.

\begin{table}[t]
  \centering
  \caption{Comparison with the hypergraph Weisfeiler--Lehman subtree kernel (HGWL), run on the same partitions with its published settings. Test scores (\%), mean $\pm$ population standard deviation over five repeats; $\Delta$ is our score minus the kernel's in the same repeat, averaged over repeats, with the numbers of repeats in which ours is higher/lower.}
  \label{tab:hgwl}
  \small
  \begin{tabular}{lccr}
    \toprule
    Dataset & Ours & HGWL & $\Delta$ \\
    \midrule
    \multicolumn{4}{l}{\emph{Macro-F1}} \\
    IMDB-Dir-Form & 59.63 $\pm$ 1.91 & 58.85 $\pm$ 3.33 & $+$0.78 (3/2) \\
    IMDB-Dir-Genre & 76.82 $\pm$ 1.37 & 74.39 $\pm$ 1.18 & $+$2.42 (4/1) \\
    Steam-Player & 62.74 $\pm$ 1.51 & 43.60 $\pm$ 2.40 & $+$19.13 (5/0) \\
    Twitter-Friend & 72.05 $\pm$ 3.68 & 54.54 $\pm$ 2.07 & $+$17.52 (5/0) \\
    RHG-10 & 98.17 $\pm$ 0.83 & 96.33 $\pm$ 0.88 & $+$1.85 (5/0) \\
    RHG-3 & 99.74 $\pm$ 0.32 & 100.00 $\pm$ 0.00 & $-$0.26 (0/2) \\
    \midrule
    \multicolumn{4}{l}{\emph{Accuracy}} \\
    IMDB-Dir-Form & 64.15 $\pm$ 2.01 & 66.60 $\pm$ 3.18 & $-$2.45 (0/5) \\
    IMDB-Dir-Genre & 80.71 $\pm$ 1.05 & 78.35 $\pm$ 0.86 & $+$2.35 (5/0) \\
    Steam-Player & 63.59 $\pm$ 1.19 & 50.78 $\pm$ 2.58 & $+$12.82 (5/0) \\
    Twitter-Friend & 74.35 $\pm$ 3.63 & 55.27 $\pm$ 2.80 & $+$19.08 (5/0) \\
    RHG-10 & 98.20 $\pm$ 0.87 & 96.40 $\pm$ 0.86 & $+$1.80 (5/0) \\
    RHG-3 & 99.73 $\pm$ 0.33 & 100.00 $\pm$ 0.00 & $-$0.27 (0/2) \\
    \bottomrule
  \end{tabular}
\end{table}

Isomorphic copies of training hypergraphs do not explain the largest margins. When isomorphism is tested without labels, no hypergraph of Steam-Player or Twitter-Friend is isomorphic to any other hypergraph of its dataset. On the other four datasets, many test hypergraphs have an isomorphic copy among the training hypergraphs. On average over the repeats, this holds for 28.4\% and 27.5\% of the test hypergraphs on IMDB-Dir-Form and IMDB-Dir-Genre, and for 78.5\% and 74.3\% on RHG-10 and RHG-3. A simple lookup predicts the most frequent class among a test hypergraph's training copies, and the training majority class for the other test hypergraphs. It reaches 31.63\%, 57.55\%, 79.76\% and 81.98\% Macro-F1 on these four datasets, which is below the method on each. Isomorphic hypergraphs receive the same features from any method that uses structure alone, so this property of the benchmark affects all the methods compared. As a check for leakage of class information, we reran the method with the training labels permuted within each repeat and everything else unchanged. Test accuracy then drops to the level of the majority-class reference and is never more than 2.8 points above it. On Twitter-Friend, Macro-F1 falls from 72.05\% to 42.96\%. On Steam-Player, the classifier predicts the majority class in every repeat.

\subsection{Overlap labels and relative frequencies (RQ2)}
\label{sec:rq2}

\begin{figure}[t]
  \centering
  \includegraphics[width=\linewidth]{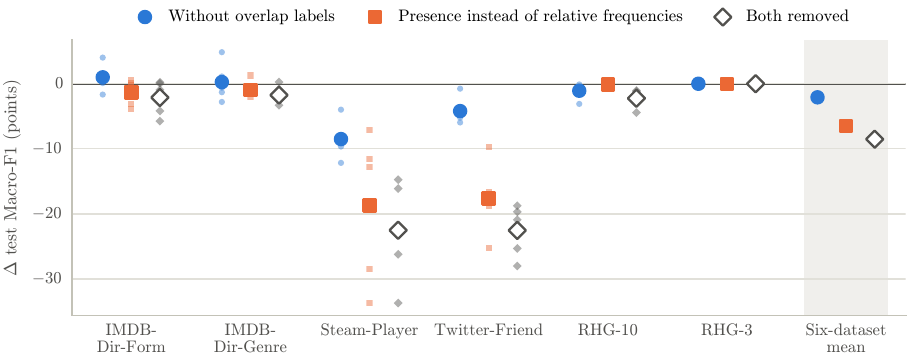}
  \caption{Overlap labels and relative frequencies. Paired difference in test Macro-F1 from the full model (points). Large markers: mean over five repeats; small markers: single repeats. The shaded group averages the dataset means over the six datasets. Table~\ref{tab:components} in Appendix~\ref{app:ablations} gives the values, the numbers of repeats higher/lower and the validation means.}
  \label{fig:components}
\end{figure}

Figure~\ref{fig:components} removes the overlap labels, replaces the relative frequencies by presence, or does both, without changing anything else. Of the two single changes, replacing relative frequencies by presence has the larger effect. Mean test Macro-F1 falls by 6.46 points, and the loss reaches 18.71 points on Steam-Player and 17.63 points on Twitter-Friend, in all five repeats on both. Validation Macro-F1 falls on these two datasets by similar amounts, 15.66 and 14.27 points, also in every repeat.

Removing the overlap labels lowers mean test Macro-F1 by 2.09 points. The loss is again largest on Steam-Player (8.50 points) and Twitter-Friend (4.20 points). The variant is also lower on RHG-10 (1.08 points), and on all three datasets it is lower in all five repeats. IMDB-Dir-Genre behaves differently. Without overlap labels, its validation Macro-F1 is 2.80 points higher in all five repeats, while its test Macro-F1 changes little (0.26 points higher, with two repeats higher and three lower). On IMDB-Dir-Form the variant is 0.99 points higher on test but 0.53 points lower on validation.

Removing both lowers mean test Macro-F1 by 8.52 points, more than either change alone, with losses of 22.51 points on Steam-Player and 22.53 points on Twitter-Friend. On the two IMDB datasets and RHG-10, presence instead of relative frequencies changes test Macro-F1 by at most 1.32 points. On RHG-3, none of the three variants changes the test scores in any repeat. Both components therefore matter most on Steam-Player and Twitter-Friend, the two datasets with the highest ratio of average hyperedges to average vertices and the highest Overlap~$\ge 2$ shares (Table~\ref{tab:datasets}).

\subsection{Pattern size, vocabulary and mining cost (RQ3)}
\label{sec:rq3}

\begin{figure}[t]
  \centering
  \includegraphics[width=\linewidth]{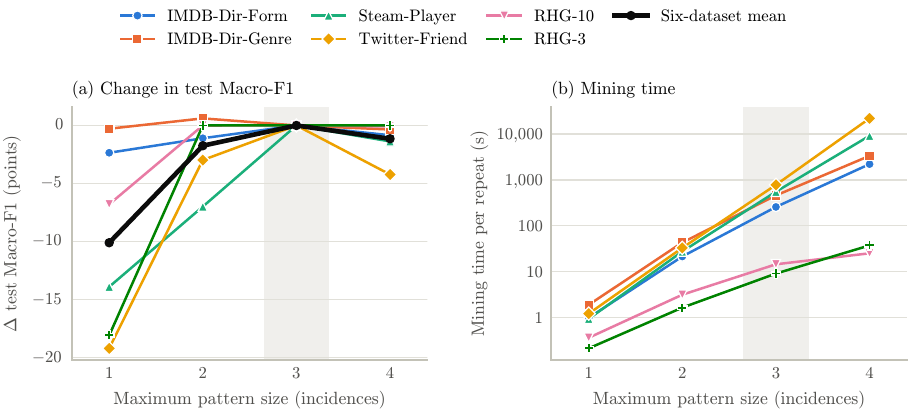}
  \caption{Maximum pattern size. (a) Difference in mean test Macro-F1 from the full model, whose maximum size is three (shaded), for each dataset and averaged over the six datasets. (b) Mean mining time per repeat on the training hypergraphs, on a logarithmic scale.}
  \label{fig:depth}
\end{figure}

Figure~\ref{fig:depth} and Table~\ref{tab:cost} vary the maximum pattern size from one to four and keep the support rule of Section~\ref{sec:support} and the classifier settings. Table~\ref{tab:repeat-counts} in Appendix~\ref{app:ablations} gives the numbers of repeats higher and lower for each size. With patterns of one incidence only, each feature is the relative frequency of one combination of a vertex-degree bin and a hyperedge-cardinality bin among the incidences of a hypergraph. This representation is 10.10 points below the full model in mean test Macro-F1, and lower in all five repeats on Steam-Player, Twitter-Friend, RHG-10 and RHG-3.

With patterns of up to two incidences, the mean difference shrinks to 1.74 points. The remaining loss is on Steam-Player (6.99 points, lower in all five repeats) and Twitter-Friend (2.98 points, lower in four). Validation agrees on both datasets (4.57 and 3.45 points, lower in all five repeats). On RHG-10 and RHG-3, size two already gives the same mean test Macro-F1 as size three. IMDB-Dir-Genre is the exception. Its Macro-F1 is higher at size two, by 0.61 points on test and by 2.18 points on validation, where it is higher in all five repeats. Patterns of at most two incidences need to occur in only two training hypergraphs, so the vocabulary at size two holds every such pattern that recurs in the training set. The patterns of three incidences are the only ones selected by the support fraction $\rho$, and their gain is concentrated on Steam-Player and Twitter-Friend.

\begin{table}[t]
  \centering
  \caption{Vocabulary size (patterns) for maximum pattern sizes 1 to 4, and the cost of the full model (size 3) per repeat: mining time on the training hypergraphs, presence-check time for the validation and test hypergraphs together (matching against the frozen vocabulary with the miner's matcher, used only to verify the counts; Section~\ref{sec:counting}), classifier training and evaluation time, and peak resident memory of the process. Means over the five repeats; the six datasets ran concurrently on one machine (Section~\ref{sec:implementation}).}
  \label{tab:cost}
  \footnotesize
  \setlength{\tabcolsep}{3pt}
  \begin{tabular}{lrrrrrrrr}
    \toprule
    & \multicolumn{4}{c}{Vocabulary size} & \multicolumn{4}{c}{Full model, per repeat} \\
    \cmidrule(lr){2-5} \cmidrule(lr){6-9}
    Dataset & Size 1 & Size 2 & Size 3 & Size 4 & Mining (s) & Presence check (s) & Classifier (s) & Memory (MB) \\
    \midrule
    IMDB-Dir-Form & 52 & 719 & 1{,}598 & 7{,}862 & 257 & 9.0 & 13.1 & 1{,}148 \\
    IMDB-Dir-Genre & 53 & 833 & 1{,}660 & 6{,}905 & 458 & 14.5 & 21.3 & 1{,}421 \\
    Steam-Player & 25 & 237 & 1{,}121 & 12{,}437 & 552 & 28.7 & 20.5 & 1{,}448 \\
    Twitter-Friend & 45 & 690 & 3{,}304 & 46{,}644 & 780 & 40.4 & 15.3 & 1{,}618 \\
    RHG-10 & 35 & 174 & 198 & 227 & 14 & 0.4 & 50.8 & 853 \\
    RHG-3 & 15 & 85 & 118 & 203 & 9 & 0.4 & 27.4 & 811 \\
    \bottomrule
  \end{tabular}
\end{table}

The cost of adding the third incidence grows faster than the vocabulary. From size two to size three, the mean mining time per repeat rises from 21--43\,s to 257--780\,s on the four real-world datasets, a factor of 11 to 24. On the synthetic datasets, it rises from 2--3\,s to 9--14\,s. Over the same step, the vocabulary grows by a factor of 2.0 to 4.8 on the real-world datasets and 1.1 to 1.4 on the synthetic ones. The presence check of the validation and test hypergraphs against the frozen vocabulary takes at most 40.4\,s per repeat, and training and evaluating the classifier at most 50.8\,s.

Patterns of up to four incidences do not raise mean test Macro-F1 on any dataset. Mean Macro-F1 is 1.14 points below the full model on test and 0.87 points below on validation. The largest loss is on Twitter-Friend (4.22 points, lower in all five repeats); on the other datasets the difference is at most 1.37 points, within the variation across classifier seeds (Section~\ref{sec:limitations}). Over the same step, the vocabulary grows by a factor of 4.2 to 14.1 on the real-world datasets (Table~\ref{tab:cost}) and the mining time by a factor of 7.3 to 28, reaching 6.1\,h per repeat on Twitter-Friend (Figure~\ref{fig:depth}).

\subsection{Sensitivity (RQ4)}
\label{sec:rq4}

\paragraph{Label resolution.} We compare exact degree, cardinality and overlap labels with the log2 bins at maximum pattern size two. Even at this size, exact labels enlarge the vocabulary 58 to 308 times on the real-world datasets, up to 212,728 patterns on Twitter-Friend (Appendix~\ref{app:ablations}). Exact labels lower mean test Macro-F1 by 3.17 points and validation Macro-F1 by 1.97 points. The largest loss is on Twitter-Friend (13.66 points, lower in all five repeats), followed by IMDB-Dir-Genre (4.85 points, all five repeats). Exact labels are higher on Steam-Player, by 1.45 points on test, but with mixed signs and within the variation across seeds. They are also higher on RHG-3, where they classify every test hypergraph correctly.

\paragraph{Support rule.} In this variant, the threshold $\rho$ applies to patterns of every size, with the phased lowering of Section~\ref{sec:support} for training hypergraphs left uncovered. Patterns of one or two incidences are then no longer kept whenever they recur in two training hypergraphs. Mean Macro-F1 falls by 0.76 points on test and 1.33 points on validation. The clearest effect is on RHG-10, where test Macro-F1 falls by 6.03 points in all five repeats (Table~\ref{tab:repeat-counts}), far more than its variation across classifier seeds (Section~\ref{sec:limitations}). On IMDB-Dir-Form, test Macro-F1 is 1.80 points higher with the uniform threshold, in all five repeats, but validation Macro-F1 is 1.23 points lower. The test difference is also smaller than the 2.11-point range of the full model's mean over five seeds.

\paragraph{Patterns of one incidence.} Removing them from the vocabulary changes mean test Macro-F1 by $+0.42$ points and validation Macro-F1 by $-0.38$ points, with mixed signs on every real-world dataset (Table~\ref{tab:repeat-counts}), so they have no consistent effect.

\paragraph{Pattern selection and classifier.} Supervised selection ranks the patterns by the entropy-based score of \citet{alam2021discriminating} on the training labels. It keeps patterns in this order until every training hypergraph contains a given number of kept patterns. For every number tested, it lowers mean test Macro-F1, from 0.67 points when the number is 60 to 11.27 points when it is 1. In the classifier, feeding the scaled relative frequencies directly to the same network instead of the learned encoder changes mean test Macro-F1 by $+0.13$ points ($-0.16$ on validation). Layer normalization and input scaling, in contrast, do matter. Without layer normalization, test Macro-F1 falls by 35.05 points on Twitter-Friend and 31.19 points on Steam-Player. Without input scaling, it falls by 30.67 points on Steam-Player. An untuned linear support vector machine with $C=1$ on the relative frequencies reaches 60.27\% Macro-F1 on Steam-Player and 71.56\% on Twitter-Friend, against 43.60\% and 54.54\% for the support vector machine with the same $C$ on the subtree kernel (Table~\ref{tab:hgwl}). Appendix~\ref{app:ablations} gives the full tables.

\subsection{Vocabulary access (RQ5)}
\label{sec:rq5}

Table~\ref{tab:vocabulary-access} lets the unlabeled validation and test hypergraphs take part in mining and leaves everything else unchanged. Mean test Macro-F1 changes by $-0.17$ points and accuracy by $-0.24$ points, with mixed signs. The largest changes are $-1.18$ points on Steam-Player (two repeats higher, three lower) and $+0.70$ points on IMDB-Dir-Form (three higher, two lower). On RHG-3 the scores do not change. The vocabulary size changes by between $-1.6\%$ and $+5.7\%$ per repeat. Access to the held-out structures during mining therefore gives no consistent benefit on these datasets.

\begin{table}[t]
  \centering
  \caption{Vocabulary access. Transductive: unlabeled validation and test hypergraphs also take part in mining; the classifier and everything else are as in the main (inductive) protocol, whose scores are in Table~\ref{tab:main-results}. Test scores (\%), mean $\pm$ population standard deviation; $\Delta$ is transductive minus inductive in the same repeat, with the numbers of repeats higher/lower. Vocabulary: range over repeats of the relative change in the size of the mined vocabulary, before patterns that occur in fewer than two training hypergraphs are dropped.}
  \label{tab:vocabulary-access}
  \small
  \setlength{\tabcolsep}{5pt}
  \begin{tabular}{lcrcrr}
    \toprule
    & \multicolumn{2}{c}{Macro-F1} & \multicolumn{2}{c}{Accuracy} & \\
    \cmidrule(lr){2-3} \cmidrule(lr){4-5}
    Dataset & Transductive & $\Delta$ & Transductive & $\Delta$ & Vocabulary (\%) \\
    \midrule
    IMDB-Dir-Form & 60.34 $\pm$ 2.22 & $+$0.70 (3/2) & 64.68 $\pm$ 2.32 & $+$0.53 (3/2) & $-$1.6 to $+$5.7 \\
    IMDB-Dir-Genre & 76.49 $\pm$ 0.71 & $-$0.32 (1/4) & 80.29 $\pm$ 0.83 & $-$0.41 (1/4) & $-$0.1 to $+$1.6 \\
    Steam-Player & 61.56 $\pm$ 2.96 & $-$1.18 (2/3) & 62.72 $\pm$ 3.03 & $-$0.87 (1/3) & $+$0.1 to $+$2.5 \\
    Twitter-Friend & 72.07 $\pm$ 3.13 & $+$0.02 (3/2) & 73.89 $\pm$ 2.62 & $-$0.46 (2/2) & $+$2.0 to $+$3.9 \\
    RHG-10 & 97.95 $\pm$ 0.89 & $-$0.22 (0/1) & 98.00 $\pm$ 0.89 & $-$0.20 (0/1) & $-$1.5 to $+$3.6 \\
    RHG-3 & 99.74 $\pm$ 0.32 & 0.00 (0/0) & 99.73 $\pm$ 0.33 & 0.00 (0/0) & $-$0.8 to 0.0 \\
    \midrule
    Mean & & $-$0.17 & & $-$0.24 & \\
    \bottomrule
  \end{tabular}
\end{table}

\section{Discussion}
\label{sec:discussion}

Relative frequencies matter most on Steam-Player and Twitter-Friend, which have the highest ratio of average hyperedges to average vertices (3.35 and 3.92, against at most 2.50 on the other datasets) and the highest Overlap~$\ge 2$ shares (Table~\ref{tab:datasets}). Nearly constant presence features explain only part of this. On Steam-Player a hypergraph contains on average 49.2\% of the vocabulary, and 15.3\% of the patterns occur in at least 90\% of the training hypergraphs, so many presence features carry little information. On Twitter-Friend a hypergraph contains 24.2\% of the vocabulary and almost no pattern is that common, yet replacing relative frequencies by presence lowers Macro-F1 by about as much (Figure~\ref{fig:components}). Presence records which configurations a hypergraph contains. The relative frequency also records how common each configuration is among the occurrences of its skeleton, and on these two datasets the ablation shows that this information carries much of the class signal. On the other four datasets a hypergraph contains 6.8\% to 15.5\% of the vocabulary, no pattern occurs in 90\% of the training hypergraphs, and relative frequencies change the scores little.

Overlap labels help on the same two datasets. On both, two hyperedges that meet at a vertex also share another vertex in about three quarters of the cases. Without overlap labels, such a pair matches the same patterns as a pair that shares a single vertex, so the vocabulary cannot separate small overlaps from large ones. IMDB-Dir-Genre prefers the vocabulary without overlap labels on validation. Overlap labels double its vocabulary, from 807 to 1,660 patterns, so each pattern is supported by fewer training hypergraphs on average. When the size of an overlap carries little class information, this may cost more than it gains. We did not test this explanation.

The gain lies in the representation rather than in the classifier. An untuned linear support vector machine on the relative frequencies stays close to the full model on Steam-Player and Twitter-Friend. It is also far above a support vector machine with the same $C$ on the subtree kernel (Section~\ref{sec:rq4}). Within the network, the learned encoder adds nothing measurable over feeding the relative frequencies directly. The DHG-Bench models also use structure alone, through a one-hot encoding of each vertex's degree, so Table~\ref{tab:main-results} compares structure-only methods. Their training and model selection differ from ours, however, so the comparison does not separate the representation from the architecture.

Most of the signal lies in patterns of at most two incidences, which the support rule keeps whenever they recur in at least two training hypergraphs. The patterns of three incidences, selected by the support fraction $\rho$, add to them on Steam-Player and Twitter-Friend, but raise the mining time 11 to 24 times on the real-world datasets. On the other four datasets, a maximum size of two gives almost the same scores. It needs under a tenth of the mining time on the IMDB datasets and about a fifth on the synthetic ones (Figure~\ref{fig:depth}), so it would be the more economical setting for them.

Letting the validation and test hypergraphs take part in mining changes the vocabulary by a few percent and the scores by less than their spread across repeats. With this access, support is counted over all hypergraphs, so held-out hypergraphs can move a pattern above or below its threshold. A pattern that occurs in fewer than two training hypergraphs is dropped before the classifier (Section~\ref{sec:cost}), so every pattern that the classifier sees occurs in at least two training hypergraphs, as in the main protocol. Access to the held-out structures therefore changes only which patterns pass the thresholds, and on these datasets this change does not alter the scores consistently.

\section{Limitations}
\label{sec:limitations}

The evidence comes from five repeated partitions of six datasets, two of which are synthetic and close to the ceiling. The comparisons are descriptive. Apart from the published DHG-Bench values, the only baseline is the hypergraph Weisfeiler--Lehman kernel, which we ran on the same partitions with its published settings. The standard deviations combine the variation between partitions with that of the classifier's initialization. We retrained the classifier with four further seeds on the same partitions and vocabularies. Within a repeat, the mean standard deviation across seeds is 1.0 to 2.1 points in test Macro-F1 on the real-world datasets and at most 0.22 points on the synthetic ones. The mean over the five repeats varies across seeds by 1.18 to 2.58 points on the real-world datasets (2.11 points on IMDB-Dir-Form) and by at most 0.14 points on the synthetic ones. Every comparison with the published results and the kernel keeps its direction under all five seeds, except two on IMDB-Dir-Form. There, the Macro-F1 difference from the kernel and the accuracy difference from the published mean change sign. The ablations use a single seed, so differences of about two points or less on the real-world datasets are within the variation that seeds alone produce. On the IMDB and RHG datasets, many test hypergraphs are isomorphic to training hypergraphs, so scores there partly reflect repeated structures.

Overlap labels describe pairs of hyperedges, so intersections of three or more hyperedges are not labeled. The log2 bins are also fixed rather than learned. The selected configuration uses patterns of at most three incidences. Size four was evaluated only with the other settings unchanged, and larger sizes were not evaluated, since mining at size four already takes up to 6.1\,h per repeat. The implementation is in Python, and the reported times were measured with the six datasets running concurrently. The method uses structure alone and does not use native vertex or hyperedge attributes.

\section{Conclusion}
\label{sec:conclusion}

Hypergraph-level classification from structure needs a fixed-length description of each hypergraph. We extended frequent sub-hypergraph mining with overlap labels. An overlap label is the binned intersection size of two pattern hyperedges, inherited from the target hypergraph. Each hypergraph is represented by the relative occurrence frequencies of the mined patterns, with a vocabulary mined from training hypergraphs only. With one configuration chosen on validation scores, the method has a higher mean Macro-F1 than the best published DHG-Bench result on five of the six datasets and is 0.21 points lower on RHG-3. Its Macro-F1 is also above that of a hypergraph Weisfeiler--Lehman kernel run on the same partitions in every repeat on Steam-Player, Twitter-Friend and RHG-10. Relative frequencies and overlap labels matter most on the two datasets with the highest ratio of average hyperedges to average vertices and the highest share of hyperedge pairs that also share a second vertex. On these datasets, patterns of three incidences also add to the smaller ones, at a higher mining cost. 

\section*{Code and data availability}

The code, the saved partitions and the configuration needed to reproduce all results are available at \url{https://github.com/NafiHasan/overlap-fhgm}. The datasets are those distributed with DHG-Bench \citep{li2025dhgbench}.

\bibliographystyle{plainnat}
\bibliography{references}

\clearpage
\appendix

\section{Hypergraph Weisfeiler--Lehman kernel variants}
\label{app:baselines}

Table~\ref{tab:baselines} reports all eight variants of the hypergraph Weisfeiler--Lehman kernels of \citet{feng2024hypergraph} that we ran on our partitions. They cross two kernels (subtree and hyperedge) with two label sets. Exact labels are the authors' input, with repeated hyperedges merged and exact degree and cardinality labels. Log2 labels are the input of our method, with every hyperedge kept and log2-binned labels. All variants use the settings of Section~\ref{sec:baselines}: four iterations, cosine normalization and a support vector machine with $C=1$. The subtree kernel with exact labels is the variant in Table~\ref{tab:hgwl}.

\begin{table}[!htb]
  \centering
  \caption{Hypergraph Weisfeiler--Lehman kernels on our partitions. Test Macro-F1 (\%), mean over five repeats; the last column is the mean over the six datasets. Exact labels: the authors' input (repeated hyperedges merged, exact degree and cardinality labels); log2 labels: the input of our method. The upper block treats colors unseen in training as in the authors' code (left out of a held-out hypergraph's norm); the lower block uses the exact cosine norm, which counts them. Our method is shown for reference.}
  \label{tab:baselines}
  \footnotesize
  \setlength{\tabcolsep}{3pt}
  \begin{tabular}{lccccccc}
    \toprule
    Variant & Form & Genre & Steam & Twitter & RHG-10 & RHG-3 & Mean \\
    \midrule
    Ours & 59.63 & 76.82 & 62.74 & 72.05 & 98.17 & 99.74 & 78.19 \\
    \midrule
    \multicolumn{8}{l}{\emph{Authors' handling of unseen colors}} \\
    Subtree, exact labels & 58.85 & 74.39 & 43.60 & 54.54 & 96.33 & 100.00 & 71.28 \\
    Subtree, log2 labels & 58.70 & 74.60 & 40.72 & 40.97 & 96.07 & 100.00 & 68.51 \\
    Hyperedge, exact labels & 58.73 & 72.90 & 53.46 & 40.78 & 89.94 & 100.00 & 69.30 \\
    Hyperedge, log2 labels & 59.26 & 75.52 & 54.83 & 53.47 & 95.76 & 100.00 & 73.14 \\
    \midrule
    \multicolumn{8}{l}{\emph{Exact cosine norm}} \\
    Subtree, exact labels & 58.59 & 75.00 & 54.17 & 42.61 & 96.25 & 100.00 & 71.10 \\
    Subtree, log2 labels & 58.43 & 74.94 & 48.58 & 37.00 & 96.49 & 100.00 & 69.24 \\
    Hyperedge, exact labels & 58.49 & 72.43 & 52.72 & 37.56 & 89.40 & 100.00 & 68.43 \\
    Hyperedge, log2 labels & 58.90 & 75.27 & 55.09 & 52.86 & 95.76 & 100.00 & 72.98 \\
    \bottomrule
  \end{tabular}
\end{table}

No variant is above our method on any dataset except RHG-3, where every kernel classifies all test hypergraphs correctly. The best kernel mean over the six datasets is 73.14\% (hyperedge kernel, log2 labels), against 78.19\% for our method. The choice of kernel and labels matters most on Steam-Player and Twitter-Friend. Under the authors' handling of unseen colors, replacing exact labels by log2 labels lowers the subtree kernel's Macro-F1 by 2.88 points on Steam-Player and 13.57 points on Twitter-Friend, whereas it raises the hyperedge kernel's by 1.37 and 12.69 points. On the other four datasets the eight variants differ by at most 3.1 points, except on RHG-10, where the hyperedge kernel with exact labels is about 6 to 7 points below the other variants. We did not select a variant on test scores; the main text uses the subtree kernel on the authors' input.

The lower block of Table~\ref{tab:baselines} uses the exact cosine norm, which also counts the colors that never occur in training. The training kernel and the fitted classifier are the same in the two blocks, so only the held-out scores change. The difference is about 0.6 points or less on the IMDB datasets and RHG-10 for the subtree kernel with exact labels, but large on the two datasets with the most distinct colors. For that variant, Macro-F1 falls from 54.54\% to 42.61\% on Twitter-Friend and rises from 43.60\% to 54.17\% on Steam-Player. For the subtree kernel with log2 labels, the Twitter-Friend value of 37.00\% equals the Macro-F1 of always predicting the training-majority class. We keep the authors' handling in the main comparison; the exact norm changes no comparison with our method.

\section{Ablation details}
\label{app:ablations}

All variants change one setting of the full model and keep the rest, including the vocabulary-size rules and the five partitions; none is reselected on validation. Differences are paired by repeat, as in Section~\ref{sec:baselines}. Table~\ref{tab:components} gives the values behind Figure~\ref{fig:components}.

\begin{table}[!htb]
  \centering
  \caption{Overlap labels and relative frequencies. Full model: test Macro-F1 (\%). Other columns: paired difference in test Macro-F1 from the full model (points), with the numbers of repeats higher/lower. The last two rows average over the six datasets, on test and on validation.}
  \label{tab:components}
  \small
  \setlength{\tabcolsep}{5pt}
  \begin{tabular}{lrrrr}
    \toprule
    Dataset & Full model & \begin{tabular}[b]{@{}r@{}}Without\\overlap labels\end{tabular} & \begin{tabular}[b]{@{}r@{}}Presence instead of\\relative frequencies\end{tabular} & \begin{tabular}[b]{@{}r@{}}Both\\removed\end{tabular} \\
    \midrule
    IMDB-Dir-Form & 59.63 & $+$0.99 (4/1) & $-$1.32 (2/3) & $-$2.11 (2/3) \\
    IMDB-Dir-Genre & 76.82 & $+$0.26 (2/3) & $-$0.98 (1/4) & $-$1.74 (1/4) \\
    Steam-Player & 62.74 & $-$8.50 (0/5) & $-$18.71 (0/5) & $-$22.51 (0/5) \\
    Twitter-Friend & 72.05 & $-$4.20 (0/5) & $-$17.63 (0/5) & $-$22.53 (0/5) \\
    RHG-10 & 98.17 & $-$1.08 (0/5) & $-$0.13 (0/1) & $-$2.23 (0/5) \\
    RHG-3 & 99.74 & 0.00 (0/0) & 0.00 (0/0) & 0.00 (0/0) \\
    \midrule
    Mean, test & 78.19 & $-$2.09 & $-$6.46 & $-$8.52 \\
    Mean, validation & 78.91 & $-$2.07 & $-$5.01 & $-$7.58 \\
    \bottomrule
  \end{tabular}
\end{table}

\paragraph{Label resolution.} Table~\ref{tab:label-resolution} compares exact and log2-binned labels at maximum pattern size two.

\begin{table}[!htb]
  \centering
  \caption{Label resolution at maximum pattern size 2, with overlap labels and relative frequencies as in the full model. Test Macro-F1 (\%), mean $\pm$ population standard deviation; $\Delta$ is exact minus log2 labels in the same repeat, on test and on validation, with the numbers of repeats higher/lower. Vocabulary: mean number of patterns.}
  \label{tab:label-resolution}
  \small
  \setlength{\tabcolsep}{4pt}
  \begin{tabular}{lccrrrr}
    \toprule
    & \multicolumn{2}{c}{Test Macro-F1} & & & \multicolumn{2}{c}{Vocabulary} \\
    \cmidrule(lr){2-3} \cmidrule(lr){6-7}
    Dataset & log2 & Exact & $\Delta$ test & $\Delta$ val. & log2 & Exact \\
    \midrule
    IMDB-Dir-Form & 58.54 $\pm$ 3.93 & 57.41 $\pm$ 2.19 & $-$1.12 (2/3) & $-$1.14 (1/4) & 719 & 41{,}842 \\
    IMDB-Dir-Genre & 77.43 $\pm$ 1.51 & 72.57 $\pm$ 1.27 & $-$4.85 (0/5) & $-$1.77 (1/4) & 833 & 66{,}252 \\
    Steam-Player & 55.75 $\pm$ 5.59 & 57.20 $\pm$ 2.08 & $+$1.45 (2/3) & $+$0.11 (3/2) & 237 & 17{,}024 \\
    Twitter-Friend & 69.07 $\pm$ 6.23 & 55.41 $\pm$ 3.70 & $-$13.66 (0/5) & $-$8.03 (0/5) & 690 & 212{,}728 \\
    RHG-10 & 98.17 $\pm$ 0.73 & 97.07 $\pm$ 1.53 & $-$1.10 (1/4) & $-$1.66 (0/5) & 174 & 798 \\
    RHG-3 & 99.74 $\pm$ 0.32 & 100.00 $\pm$ 0.00 & $+$0.26 (2/0) & $+$0.64 (4/0) & 85 & 721 \\
    \midrule
    Mean & & & $-$3.17 & $-$1.97 & & \\
    \bottomrule
  \end{tabular}
\end{table}

\paragraph{Other ablations.} Table~\ref{tab:ablations} gives the paired differences in test Macro-F1 on each dataset, and the mean differences on test and validation, for the remaining ablations.

\begin{table}[!htb]
  \centering
  \caption{One-factor ablations around the full model. Paired difference in test Macro-F1 from the full model (points; mean over five repeats) on each dataset, and its mean over the six datasets on test and on validation. Selection keeps patterns ranked by the entropy-based score until every training hypergraph contains at least the target number of kept patterns.}
  \label{tab:ablations}
  \footnotesize
  \setlength{\tabcolsep}{3pt}
  \begin{tabular}{lrrrrrrrr}
    \toprule
    Variant & Form & Genre & Steam & Twitter & RHG-10 & RHG-3 & Test & Val. \\
    & \multicolumn{6}{c}{$\Delta$ test Macro-F1} & \multicolumn{2}{c}{$\Delta$ mean} \\
    \cmidrule(lr){2-7} \cmidrule(lr){8-9}
    \multicolumn{9}{l}{\emph{Support rule and one-incidence patterns}} \\
    Uniform support & $+$1.80 & $-$1.32 & $+$0.11 & $+$0.89 & $-$6.03 & 0.00 & $-$0.76 & $-$1.33 \\
    Without patterns of one incidence & $+$0.56 & $-$0.09 & $-$0.03 & $+$2.11 & $-$0.02 & 0.00 & $+$0.42 & $-$0.38 \\
    \midrule
    \multicolumn{9}{l}{\emph{Supervised pattern selection (coverage target)}} \\
    Selection, target 60 & $+$0.47 & $+$0.42 & $-$1.82 & $-$3.06 & 0.00 & 0.00 & $-$0.67 & $-$0.76 \\
    Selection, target 30 & $+$0.22 & $+$1.14 & $-$3.05 & $-$7.67 & $-$0.17 & 0.00 & $-$1.59 & $-$1.85 \\
    Selection, target 15 & $+$0.40 & $+$0.21 & $-$3.05 & $-$9.45 & $-$0.02 & 0.00 & $-$1.98 & $-$2.09 \\
    Selection, target 5 & $-$0.03 & $-$0.85 & $-$7.37 & $-$14.07 & $-$0.13 & 0.00 & $-$3.74 & $-$3.52 \\
    Selection, target 1 & $-$2.58 & $-$1.64 & $-$8.52 & $-$15.79 & $-$11.99 & $-$27.11 & $-$11.27 & $-$10.31 \\
    \midrule
    \multicolumn{9}{l}{\emph{Classifier}} \\
    No encoder, frequencies into the head & $+$1.85 & $-$3.01 & $+$0.52 & $+$1.28 & $+$0.13 & 0.00 & $+$0.13 & $-$0.16 \\
    No encoder, linear SVM & $-$1.02 & $-$1.89 & $-$2.46 & $-$0.49 & $-$0.18 & 0.00 & $-$1.01 & $-$1.43 \\
    Without the log-count input & $+$0.79 & $-$1.01 & $-$0.87 & $-$0.71 & $-$0.19 & 0.00 & $-$0.33 & $+$0.31 \\
    Unscaled relative frequencies & $+$0.34 & $+$0.59 & $-$30.67 & $-$2.23 & $-$0.02 & 0.00 & $-$5.33 & $-$5.32 \\
    Divided by training standard deviation & $+$0.66 & $-$0.86 & $-$1.64 & $-$2.03 & $-$0.09 & 0.00 & $-$0.66 & $-$1.29 \\
    One hidden layer (64) & $+$0.94 & $-$1.52 & $+$0.11 & $-$1.08 & $-$0.09 & 0.00 & $-$0.27 & $-$0.08 \\
    Without layer normalization & $-$13.54 & $-$12.44 & $-$31.19 & $-$35.05 & $-$2.53 & 0.00 & $-$15.79 & $-$16.06 \\
    Pattern dropout 0 (instead of 0.3) & $+$1.52 & $-$0.56 & $+$0.79 & $-$1.34 & $-$0.12 & 0.00 & $+$0.05 & $+$0.06 \\
    Encoder L2 0.01 (instead of 0.05) & $+$0.05 & $-$0.84 & $-$0.10 & $-$1.72 & $-$0.09 & 0.00 & $-$0.45 & $-$0.46 \\
    Embedding dimension 8 & $-$0.16 & $-$0.15 & $-$0.40 & $-$0.72 & $-$0.09 & 0.00 & $-$0.25 & $+$0.41 \\
    Embedding dimension 16 & $-$1.19 & $-$0.06 & $-$0.09 & $-$1.83 & $-$0.09 & 0.00 & $-$0.54 & $+$0.11 \\
    Embedding dimension 64 & $+$0.78 & $-$0.47 & $-$0.71 & $+$0.10 & 0.00 & 0.00 & $-$0.05 & $-$0.50 \\
    \bottomrule
  \end{tabular}
\end{table}

The support rule and the patterns of one incidence matter little on average (Table~\ref{tab:ablations}, first block). Uniform support lowers the mean test Macro-F1 by 0.76 points, mainly through RHG-10 ($-6.03$); IMDB-Dir-Form gains 1.80 points on test but loses 1.23 on validation. Removing the patterns of one incidence changes the test mean by $+0.42$ points and the validation mean by $-0.38$ points.

Supervised selection lowers the mean at every coverage target tested, from $-0.67$ points at a target of 60 to $-11.27$ at a target of 1. The losses are concentrated on Steam-Player and Twitter-Friend; a target of 1 also fails on RHG-3 ($-27.11$ points), where the classifier is otherwise exact. Keeping all patterns that pass the support rule, as the full model does, was never worse on average than any selection we tried.

In the classifier, the learned encoder is not essential: feeding the scaled relative frequencies directly to the network changes the test mean by $+0.13$ points and the validation mean by $-0.16$ points, and a linear support vector machine on the same input is 1.01 points lower on test. Input scaling and layer normalization matter. Without layer normalization the test mean falls by 15.79 points, by 35.05 on Twitter-Friend and 31.19 on Steam-Player. Unscaled relative frequencies lose 30.67 points on Steam-Player. The hidden layer, pattern dropout, the encoder penalty and the embedding dimension each change the test mean by less than 0.6 points.

\paragraph{Repeat counts.} Table~\ref{tab:repeat-counts} gives the numbers of repeats higher and lower, on test and on validation, for the pattern-size and support ablations whose repeat-level directions are discussed in Sections~\ref{sec:rq3} and~\ref{sec:rq4}.

\begin{table}[!htb]
  \centering
  \caption{Repeat-level view of the pattern-size and support ablations. Paired difference in Macro-F1 from the full model (points; mean over five repeats), with the numbers of repeats higher/lower, on test and on validation. Size 3 is the full model.}
  \label{tab:repeat-counts}
  \small
  \setlength{\tabcolsep}{5pt}
  \begin{tabular}{lrrrrr}
    \toprule
    Dataset & \begin{tabular}[b]{@{}r@{}}Maximum\\size 1\end{tabular} & \begin{tabular}[b]{@{}r@{}}Maximum\\size 2\end{tabular} & \begin{tabular}[b]{@{}r@{}}Maximum\\size 4\end{tabular} & \begin{tabular}[b]{@{}r@{}}Uniform\\support\end{tabular} & \begin{tabular}[b]{@{}r@{}}No one-\\incidence\\patterns\end{tabular} \\
    \midrule
    \multicolumn{6}{l}{\emph{Test}} \\
    IMDB-Dir-Form & $-$2.35 (1/4) & $-$1.10 (1/4) & $-$0.84 (2/3) & $+$1.80 (5/0) & $+$0.56 (3/2) \\
    IMDB-Dir-Genre & $-$0.28 (1/4) & $+$0.61 (3/2) & $-$0.35 (3/2) & $-$1.32 (1/4) & $-$0.09 (2/3) \\
    Steam-Player & $-$13.89 (0/5) & $-$6.99 (0/5) & $-$1.37 (2/3) & $+$0.11 (3/2) & $-$0.03 (3/2) \\
    Twitter-Friend & $-$19.19 (0/5) & $-$2.98 (1/4) & $-$4.22 (0/5) & $+$0.89 (3/2) & $+$2.11 (3/2) \\
    RHG-10 & $-$6.79 (0/5) & 0.00 (1/1) & $-$0.09 (0/1) & $-$6.03 (0/5) & $-$0.02 (1/1) \\
    RHG-3 & $-$18.08 (0/5) & 0.00 (0/0) & 0.00 (0/0) & 0.00 (0/0) & 0.00 (0/0) \\
    \midrule
    \multicolumn{6}{l}{\emph{Validation}} \\
    IMDB-Dir-Form & $-$1.69 (1/4) & $-$1.69 (1/4) & $-$1.54 (1/4) & $-$1.23 (1/4) & $-$0.73 (1/4) \\
    IMDB-Dir-Genre & $+$2.15 (3/2) & $+$2.18 (5/0) & $+$1.04 (5/0) & $+$1.77 (5/0) & $+$1.92 (5/0) \\
    Steam-Player & $-$15.01 (0/5) & $-$4.57 (0/5) & $-$2.83 (3/2) & $-$1.97 (0/5) & $-$2.23 (0/5) \\
    Twitter-Friend & $-$15.37 (0/5) & $-$3.45 (0/5) & $-$1.64 (2/3) & $-$0.43 (2/3) & $-$0.87 (1/4) \\
    RHG-10 & $-$6.79 (0/5) & $-$0.26 (1/1) & $-$0.27 (0/1) & $-$6.13 (0/5) & $-$0.39 (0/2) \\
    RHG-3 & $-$20.66 (0/5) & 0.00 (0/0) & 0.00 (0/0) & 0.00 (0/0) & 0.00 (0/0) \\
    \bottomrule
  \end{tabular}
\end{table}

\end{document}